\documentclass[journal]{IEEEtran}

\ifCLASSINFOpdf
\else
   \usepackage[dvips]{graphicx}
\fi
\usepackage{url}

\usepackage{graphicx}
\usepackage{booktabs}
\usepackage{multirow}
\usepackage{makecell}
\usepackage{amssymb}
\usepackage{subfigure}
\usepackage{amssymb}
\usepackage[noend]{algpseudocode} 
\usepackage{algorithmicx,algorithm}
\usepackage{cite}
\usepackage{balance}

\begin{document}

\title{Sub-action Prototype Learning for Point-level Weakly-supervised Temporal Action Localization}


\author{Yueyang Li, Yonghong Hou, Wanqing Li


}

\markboth{Journal of \LaTeX\ Class Files}
{Shell \MakeLowercase{\textit{et al.}}: Bare Demo of IEEEtran.cls for IEEE Journals}
\maketitle

\begin{abstract}
Point-level weakly-supervised temporal action localization (PWTAL) aims to localize actions with only a single timestamp annotation for each action instance. Existing methods tend to mine dense pseudo labels to alleviate the label sparsity, but overlook the potential sub-action temporal structures, resulting in inferior performance. To tackle this problem, we propose a novel sub-action prototype learning framework (SPL-Loc) which comprises Sub-action Prototype Clustering (SPC) and Ordered Prototype Alignment (OPA). SPC adaptively extracts representative sub-action prototypes which are capable to perceive the temporal scale and spatial content variation of action instances. OPA selects relevant prototypes to provide completeness clue for pseudo label generation by applying a temporal alignment loss. As a result, pseudo labels are derived from alignment results to improve action boundary prediction. Extensive experiments on three popular benchmarks demonstrate that the proposed SPL-Loc significantly outperforms existing SOTA PWTAL methods. 
\end{abstract}

\begin{IEEEkeywords}
Weakly-supervised temporal action localization, Point-level supervision, Sub-action prototype learning
\end{IEEEkeywords}

\IEEEpeerreviewmaketitle

\section{Introduction}

\IEEEPARstart{T}{emporal} action localization (TAL) is an important visual task with numerous applications (e.g., anomaly detection \cite{2} and video retrieval \cite{3}) and has witnessed remarkable progress in the fully-supervised setting \cite{4}. To bypass the tedious manual annotations of action boundaries, video-level weakly-supervised TAL methods \cite{5,6,7} has draw increasing attention which localizes actions with only video-level class labels. However, due to the absent of explicit location supervision, they suffer from action-background confusion and drop largely behind the fully-supervised methods.

To balance annotation costs and model performance, point-level weakly-supervised temporal action localization (PWTAL) is proposed, where only one single timestamp (point) is annotated within each action instance for training. To date in the literature, pioneering PWTAL methods divide the input video into a series of snippets and seek to generate dense pseudo labels to provide snippet-level supervision. SF-Net \cite{8} mines potential action and background frames to expand point annoations. Ju et al. \cite{9} design a differentiable mask generator. LACP \cite{10} proposes a greedy algorithm to search for proposals with high confidences as fine-grained supervision. Nevertheless, these methods handle the information within each snippet independently, overlooking the potential temporal structures of action instances, thus can only obtain suboptimal pseudo labels and fail to learn action completeness.

\begin{figure}
\centerline{\includegraphics[width=1\columnwidth]{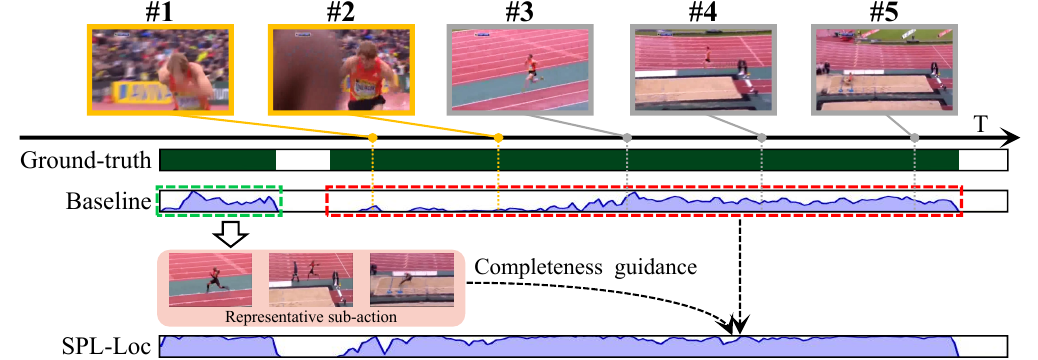}}
\setlength{\abovecaptionskip}{-1pt}
\caption{Comparison between localization results of baseline and SPL-Loc. The blue dashed box denotes a representative instance and the green one denotes an outlier instance. The outlier instance is hard to be detected by baseline, because camera view has changed and only part of the athlete's body appears in the snippet \#1 and \#2. By contrast, the representative instance with less interference and the entire body is much easier to be detected.}
\end{figure}

To illustrate the above issue, we take {\em “LongJump”} in Fig. 1 as an example. Baseline is a conventional method that only performs at snippet-level. It completely capture the representative instance (e.g., the blue dashed box) but fail to detect the outlier instance (e.g., the green dashed box) with camera view change. One common detail is that a class of actions is generally involves multiple sub-actions that occur sequentially, i.e., sub-action level motion pattern, no matter how the appearance changes. Although the two action instances differ in appearance, they both consists of three ordered sub-actions: {\em “approach run”} (snippet \#1, \#2 and \#3), {\em “takeoff”} (snippet \#4) and {\em “landing”} (snippet \#5). Accordingly, we argue that the sub-action temporal dependency of representative instances can serve as a bridge to offer completeness guidance for outlier instances, which helps thoroughly explore intrinsic motion pattern of a certain action class.

This observation motivate us to exploit the potential completeness clues, i.e., representative sub-action temporal dependency, to mine intrinsic motion pattern at sub-action level, so as to promote pesudo label generation. A primary challenge is to obtain sub-actions features from representative instances with large temporal scale variation and apply their temporal dependency to guide other instances. To this end, take advantage of prototypical features' noise robustness \cite {14}, we propose a novel \textbf{S}ub-action \textbf{P}rototype \textbf{L}earning framework for point-level weakly-supervised temporal action \textbf{Loc}alization (SPL-Loc), which contains two modules: Sub-action Prototype Clustering (SPC) and Ordered Prototype Alignment (OPA). SPC does sub-action prototype extraction from representative proposals and adaptively change the prototype count. This makes the extracted prototypes temporal-aware and spatially-adapted, so as to be robust enough to handle the variations caused by video noise such as camera view change, incomplete body and motion ambiguous. 
For the partition of the input video into multiple region, OPA selects proposal-related prototypes to perform temporal alignment \cite{15} with undeterminated region, which helps deliver sub-action level completeness cues among different instances. Furthermore, the alignment results can serve as online pseudo labels to provide fine-grained supervision. The main contributions can be summarized as follows: 1) A unified pseudo label generation framework is designed to mine intrinsic motion pattern at sub-action level for PWTAL. 2) Two modules, i.e., SPC and OPA, are proposed to extract sub-action prototypes and provide completeness guidance during pesudo label generation respectively. 3) The experimental results highlight the benefits of our method, which significantly outperforms SOTA methods.

\section{Methodology}

\subsection{Baseline Setup}

As shown in Fig. 2, given an $T$-snippets input video, a set of point annoations is provided $\{ (y_{{t_i}}^{act},{t_i})\} _{i = 1}^{{N^{act}}}$ for all ${N^{act}}$ instances, where ${t_i}$ and $y_{{t_i}}^{act}$ provide location and class supervision respectively. A feature extractor is employed to encode motion (FLOW) and appearance (RGB) information of the input video. Both RGB and FLOW features are fed in two convolutional layers to learn $D$-dimensional task-specific embedding ${X}\in {{\mathbb{R}}^{{T}\times D}}$. Subsequently, temporal class activation sequence (TCAS) ${\mathcal A}\in {{\mathbb{R}}^{{T}\times (C+1)}}$, representing the snippet-level action scores, is derived from another two convolutional layers with sigmoid and average operation. The $(C+1)$-$th$ dimension corresponds to the background score. Our baseline introduces video- and point-level loss for network training. For each class, we perform the $top$-$k$ pooling operation \cite{16} to produce video-level class score which is supervised by a cross-entropy loss $\mathcal{L}_{video}$. In addition, following \cite{10}, we search for pseudo background points between any two adjacent action points by applying a threshold on the background score to supplement the point annotations. A focal loss $\mathcal{L}_{point}$ \cite{17} is adopted to supervise the action points and the pseudo background points. The total loss of baseline is as follows:\begin{equation}
{\mathcal L_{base}} = {\lambda _1}{\mathcal L_{video}} + {\lambda _2}{\mathcal L_{point}}
\end{equation} where ${\lambda_*}$ is the hyper-parameter to balance the loss terms.

\begin{figure*}
\centerline{\includegraphics[width=0.8\linewidth]{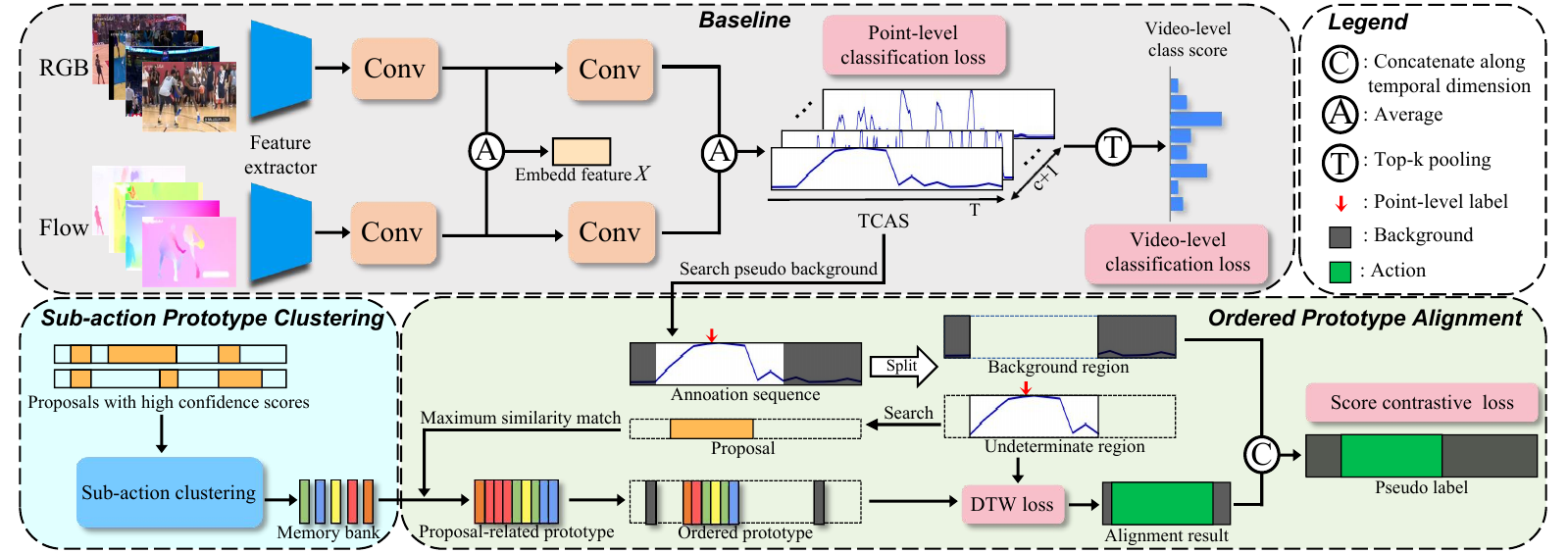}}
\setlength{\abovecaptionskip}{-1pt}
\caption{Overall architecture of SPL-Loc, which contains two modules: Sub-action Prototype Clustering (SPC) and Ordered Prototype Alignment (OPA).}
\end{figure*}

\subsection{Sub-action Prototype Learning}

Although the pseudo background points can supplement the sparse point annoations to some extent, baseline still requires dense pesudo-labels and suffers from incomplete predictions. Nevertheless, the proposals with high confidences produced by baseline can still provide a decent approximation of action durations. To take full advantage of this prior knowledge, we regard the proposals with high confidences as representative propoals and introduce Sub-action Prototype Clustering (SPC) and Ordered Prototype Alignment (OPA).

\textbf{Sub-action Prototype Clustering.} Recent work \cite{18} construct class prototypes for snippet-level relationship modeling but merely use a single prototype to represent a whole action instance, which drop temporal scale information unavoidably. Intuitively, an action instance with longer duration need more prototypes to capture the variations of temporal scale and spatial content. Consequently, SPC focus on a multiple prototypes learning strategy to adaptively change the prototype count according to the proposal duration. Formally, given a representative proposal ${{X}^{p}}\in {{\mathbb{R}}^{{{N}_{p}}\times D}}$ and uniformly initialized sub-action prototypes ${{S}^{0}}\in {{\mathbb{R}}^{{{N}_{s}}\times D}}$ within ${{X}^{p}}$, where ${N}_{p}$ is the proposal length and ${N}_{s}$ denotes the number of prototypes. We define the distance metric $Dis$ by calculating the feature similarity while also considering the temporal similarity. \begin{equation}
Dis=\sqrt{{{({{d}_{f}})}^{2}}+\gamma {{({{d}_{t}})}^{2}}}
\end{equation} where ${d}_{t}$ and ${d}_{f}$denote the Euclidean distance for temporal position and features respectively. $\gamma >0$ is a trade-off parameters. Afterward, we extract sub-action prototypes by iterative computing association and weighted update. The process of SPC is delineated in Algorithm 1.

\begin{algorithm}[!t]\scriptsize 
\caption{Sub-action Prototype Clustering (SPC)}
\hspace*{0.0in} {\bf Input:} 
proposal feature ${{X}^{p}}$, initial sub-action prototypes ${{S}^{0}}$ \\
\hspace*{0.0in} {\bf Output:} 
final sub-action prototypes ${{S}^{\mu }}$
\begin{algorithmic}[1]
\For {$\mu$ in $\{1, 2, ..., L\}$}
\State Compute distance function, $Dis=\sqrt{{{({{d}_{f}})}^{2}}+\gamma {{({{d}_{t}})}^{2}}}$
\State Compute association between each snippet $X_{i}^{p}$ and sub-action prototype $S_{j}^{\mu -1}$, $A_{ij}^{\mu }=-{{e}^{-Dis(X_{i}^{p},S_{j}^{\mu -1})}}$
\State Update sub-action prototypes, $S_{i}^{\mu }=\frac{1}{\sum\nolimits_{i}{A_{ij}^{\mu }}}\sum\limits_{i}^{{{N}_{P}}}{A_{ij}^{\mu }X_{i}^{p}}$
\EndFor
\State \Return final sub-action prototypes ${{S}^{\mu }}$
\end{algorithmic}
\end{algorithm}

As discussed above, the main contribution of SPC is its adaptive perception to temporal scale and spatial content. To guarantee this adaptability, we constrain the number of prototypes according to the following adaptive criterion. \begin{equation}
{{N}_{s}}=\min [(\frac{{{N}_{p}}}{{{r}_{p}}}),{{N}_{max}}]
\end{equation} where ${{r}_{p}}$ indicates the average length corresponding to each initial sub-action prototype. To prevent sub-actions from being oversegmented, we set hyper-parameter ${{N}_{max}}$ to constrain the upper limit of ${{N}_{s}}$. After the prototype extraction, a memory bank ${\mathcal{M}^{sub}}$ is employed to store sub-action prototypes for each class which enables us to mine intrinsic motion pattern in an inter-video fashion. For each training epoch, only the prototypes of the $top$-${k}_{sub}$ confidences are retained in ${\mathcal{M}^{sub}}$.

\textbf{Ordered Prototype Alignment.} As shown in Fig. 2, based on the annotation sequence, the input video can be split into ${N^{act}}$ undeterminated regions $\{{X_n^{un}}\} _{n = 1}^{{N^{act}}}$ and ${N^{act}}+1$ background regions $\{{X_n^{bkg}}\} _{n = 1}^{{N^{act}+1}}$. Each undeterminated regions contains an action instance. Different from \cite{10} which directly searches proposals within undeterminate regions as pseudo-labels, OPA aims to refine video features while generating pseudo-labels capable of covering more complete actions by introducing dynamic time warping (DTW) \cite{19}.

Given an $M_n$-snippets undeterminated region feature ${X_n^{un}}\in {{\mathbb{R}}^{{M_n}\times D}}$ and the $N_n$-snippets proposal feature ${X_n^{pro}}\in {{\mathbb{R}}^{{N_n}\times D}}$ searched within it. We measure the similarity between each proposal snippet and each prototype in ${\mathcal{M}^{sub}}$ of the same class by calculating the cosine distance. The proposal-related prototype sequence ${P_n^{sub}}\in {{\mathbb{R}}^{{N_n}\times D}}$ is generated by selecting the prototypes with the maximum similarity to the proposal snippets in order. The consecutively repeated prototypes are dropped to reduce computational redundancy. Assuming that undeterminated regions evolve in a background-action-background order, we temporally average pooling the background regions ${X_n^{bkg}}$ and ${X_{n+1}^{bkg}}$ adjacent to ${X_n^{un}}$ to produce the background prototypes${P_n^{bkg}}\in {{\mathbb{R}}^{{1}\times D}}$ and ${P_{n+1}^{bkg}}\in {{\mathbb{R}}^{{1}\times D}}$ respectively. Subsequently, we obtain the ordered prototype sequence, carrying the completeness guidance information of ${X_n^{un}}$,  by concatenation: ${P_n^{ord}}= Concate({P_{n}^{bkg}}, {P_n^{sub}}, {P_{n+1}^{bkg}}) \in {{\mathbb{R}}^{{(N_n+2)}\times D}}$. The cumulative distance function between the ordered prototype sequence ${P_n^{ord}}$ and the undeterminated region ${X_n^{un}}$ is evaluated with the following recursion: \begin{equation}
S(i,j)=Cos(i,j)+min\{S(i,j-1), S(i-1,j-1)\}
\end{equation} where $Cos(\cdot)$ represents cosine distance, $S(i,j)$ is evaluated on the $i$-$th$ prototype of ${P_n^{ord}}$ and the $j$-$th$ snippet of ${X_n^{un}}$. Notely, different from conventional DTW methods, we impose rigid constraints on warping paths to guarantee that each snippet can only be aligned to a single prototype. Finally, pesudo label is derived from the combination of alignment results for all undeterminated regions. The snippets aligned with the sub-action prototypes are considered as positive labels, while those aligned with the background prototypes are considered as negative labels

After the above recursion, the distance measure between ${P_n^{ord}}$ and ${X_n^{un}}$ is provided by: ${\phi}(P_n^{ord},X_n^{un})=S(M_n,N_n+2)$. It is evident that the  undeterminated region $X_n^{un}$ should be more similar to the ordered prototype of the same class $P_n^{ord}$ than the ordered prototype of different class $\bar P_{n;c}^{ord}$. To this end, we design the following contrasting loss: \begin{equation}
{\mathcal L_{OPA}} =  - \frac{1}{{{N^{act}}}}\sum\limits_{n = 1}^{{N^{act}}} {\log [\frac{{\exp (-\phi(P_n^{ord},X_n^{un})/\tau )}}{{\sum\nolimits_{c = 1}^C {\exp (-\phi(\bar P_{n;c}^{ord},X_n^{un})/\tau )}}}]}\end{equation} where $\tau$ denotes the temperature parameter.

\textbf{Training Objective.} Following \cite{10}, score contrastive loss ${\mathcal L_{PL}}$ is introduced to provide fine-grained supervision through the pseudo labels. The total loss is composed as follows:
\begin{equation}
{\mathcal L_{total}} = {\mathcal L_{base}} + {\lambda _3}{\mathcal L_{OPA}} + {\lambda _4}{\mathcal L_{PL}}
\end{equation} where ${\lambda_*}$ is the hyper-parameter to balance the loss terms.

\subsection{Inference}

During inference, we apply the multi-threshold approach \cite{multi-threshold-TCAS} on TCAS to get proposals and choose outer-inner-contrast score \cite{score} to calculate the confidence for each proposal. Finally, NMS \cite{NMS} is performed to remove redundant proposals. Note that SPC and OPA are not performed at testing time.

\section{Experiments}

\subsection{Datasets}

We evaluate SPL-Loc on three temporal action localization datasets. THUMOS-14 \cite{22} contains 200 validation and 213 test videos, which belong to 20 action categories. The video length varies over a wide range, making this dataset extremely challenging. GTEA \cite{23} consisting of 7 fine-grained daily actions in kitchen, has 21 training videos and 7 test videos. In BEOID \cite{24}, there are 58 videos from 30 action categories.

\subsection{Implementation Details}

In accordance with previous PWTAL methods \cite{8,9,10}, we report mean average precision (mAP) at different IoU thresholds to evaluate performance. I3D \cite{25} is employed to extract video features, which is not fine-tuned for fair comparison. The feature dimension $D$ is set to 1024. For all three datasets, we set ${\lambda_1}=1$, ${\lambda_2}=1$, ${\lambda_3}=1.5$, ${\lambda_4}=1$. The number of iterations $L$ is set to 6. For THUMOS-14, we set ${{r}_{p}}=5$, $\gamma=3$, ${k}_{sub}=10$. For GTEA and BEOID we set ${{r}_{p}}=3$, $\gamma=1$, ${k}_{sub}=8$.

\subsection{Ablation studies}

To evaluate the effectiveness of SPL-Loc, we conduct experiment with ablation on THUMOS-14.

{\em 1) The impact of the adaptive criterion:} Table I reports the impact of the maximum number of prototype ${N}_{max}$. Notably, SPC degenerates to global average pooling when ${N}_{max}=1$. As ${N}_{max}$ increases, the perception ability of the prototype to temporal scale is gradually enhanced. When ${N}_{max}=5$, SPL-Loc achieves the optimal performance and does not improve with the increase of ${N}_{max}$ after that. We infer that it leads to over-segmentation for some representative actions when ${N}_{max}$ is too large, which does not further help to learn action completeness. We also compared our adaptive criterion with fixed criterion (${N}_{max}=5$) in Table II. Even with the fixed number of prototypes, our approach still exceed the method that generates generates pseudo labels using only proposals. Besides, introducing the adaptive criterion can generate better pseudo labels while reduce computational redundancy.

\begin{table}[!t]\scriptsize 
\caption{Ablations on the maximum number of sub-action prototypes ${N}_{max}$. AVG is the average mAP at IoU threshold 0.3:0.1:0.7.}
\renewcommand{\arraystretch}{1}
\centering
\begin{tabular}{c|ccccc|c}
\toprule
\multirow {2}*{${N}_{max}$} & \multicolumn {5}{c|}{mAP@IoU(\%)} & \multirow {2}*{AVG} \\
& 0.3 & 0.4 & 0.5 & 0.6 & 0.7 & \\
\hline
1 & 64.8 & 56.8 & 45.7 & 34.4 & 21.2 & 44.6\\
2 & 65.8 & 58.0 & 45.6 & 34.9 & 21.5 & 45.2\\
3 & 66.3 & 57.8 & 45.9 & 35.3 & 22.0 & 45.5\\
4 & 66.3 & 58.4 & 46.5 & 35.8 & 22.4 & 45.9\\
5 & \textbf{66.9} & 58.4 & \textbf{46.8} & \textbf{36.3} & \textbf{22.9} & \textbf{46.3}\\
6 & 66.7 & \textbf{58.5} & 46.4 & 36.0 & 22.7 & 46.1\\
\bottomrule
\end{tabular}
\end{table}

\begin{table}[!t]\scriptsize 
\caption{Comparison between fixed and adaptive criterion}
\renewcommand{\arraystretch}{1}
\centering
\begin{tabular}{c|ccccc|c}
\toprule
\multirow {2}*{Criterion} & \multicolumn {5}{c|}{mAP@IoU(\%)} & \multirow {2}*{AVG} \\
& 0.3 & 0.4 & 0.5 & 0.6 & 0.7 & \\
\hline
Only proposal & 65.3 & 57.0 & 45.8 & 34.2 & 21.4 & 44.7\\
Fixed & 66.3 & 58.2 & 46.1 & 35.5 & 22.1 & 45.6\\
Adaptive & \textbf{66.9} & \textbf{58.4} & \textbf{46.8} & \textbf{36.3} & \textbf{22.9} & \textbf{46.3}\\
\bottomrule
\end{tabular}
\end{table}

{\em 2) The contribution of each component:} As shown in Table III, the upper section reports the effect of the proposed method on baseline. Even though without pseudo label, our model (id 3) still achieves absolute gain of 1.8\% in terms of average mAP. This clearly shows that the proposed OPA is able to refine the temporal context of action instances, leading to better action-background separation and better inter-class distinction. The lower section reports the effect of the proposed method on pseudo label generation. Especially, in experiment 5, we wipe out the SPC and pick the same number of representative snippets to perform temporal alignment at snippet-level for fair comparison. This setting achieves only a slight improvement of 0.1\%, much less than 1.5\%. This is because our sub-action prototypes are able to perceive the temporal scale and spatial content and produce better pseudo labels than representative snippets. The similar phenomenon also appears in the upper section. Therefore, we argue that mine intrinsic motion pattern at sub-action level is more important than snippet-level.

\begin{table}[!t]\scriptsize 
\caption{Contribution of each component.}
\centering
\begin{tabular}{c|cccc|ccc|c}
\toprule
\multirow {2}*{ID} & \multirow {2}*{$\mathcal{L}_{base}$} & \multirow {2}*{SPC} & \multirow {2}*{$\mathcal{L}_{OPA}$} & \multirow {2}*{$\mathcal{L}_{PL}$} & \multicolumn {3}{c|}{mAP@IoU(\%)} & \multirow {2}*{AVG} \\
& & & & & 0.3 & 0.5 & 0.7 & \\
\hline
1 & $\checkmark$ & & & & 61.3 & 41.9 & 15.9 & 40.4 \\
2 & $\checkmark$ & & $\checkmark$ & & 61.0 & 42.4 & 16.1 & 40.7 \\
3 & $\checkmark$ & $\checkmark$ & $\checkmark$ & & 63.2 & 43.7 & 17.2 & 42.2 \\
\hline
4 & $\checkmark$ & & & $\checkmark$ & 65.3 & 45.8 & 21.4 & 44.7 \\
5 & $\checkmark$ & & $\checkmark$ & $\checkmark$ & 65.8 & 45.6 & 21.0 & 44.8 \\
6 & $\checkmark$ & $\checkmark$ & $\checkmark$ & $\checkmark$ & \textbf{66.9} & \textbf{46.8} & \textbf{22.9} & \textbf{46.3}\\
\bottomrule
\end{tabular}
\end{table}

\subsection{Qualitative results}

As shown in Fig. 3. the method that generate pseudo labels using only proposals can produces polarized TCAS scores but still fails to localize complete actions. In contrast, our SPL-Loc can produce more accurate pseudo labels and more complete localization results regardless of the video noise.

\begin{figure} 
\centering  
\setlength{\abovecaptionskip}{-1pt}
\subfigcapskip=-3pt
\subfigure[ An example of {\em “ThrowDiscus”}]{
\includegraphics[width=0.8\columnwidth]{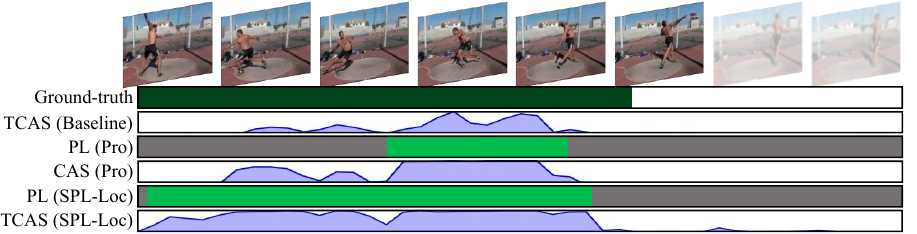}}
\subfigure[An example of {\em “HighJump”}]{
\includegraphics[width=0.8\columnwidth]{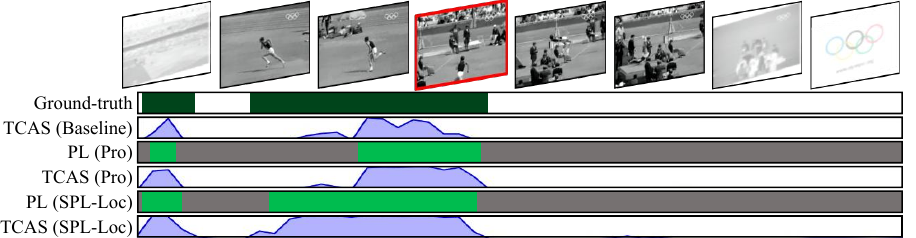}}
\caption{Visualization of pseudo labels (PL) and CAS. “Pro” denotes only proposals are used to generate pseudo labels. (a) is a slow long action. (b) has a long “approach run” and part of the athlete's body is missing when a camera view change occurs (highlighted with red box).}
\end{figure}

\subsection{Comparison with SOTA methods}

As shown in Table III, in both manual labels \cite{8} and uniform distribution labels \cite{26}, SPL-Loc surpasses the best competitor LACP, showing its robustness to label selection. Meanwhile, SPL-Loc exceeds video-level methods by a large margin with affordable annotations cost. Table IV summarize the results on GTEA and BEOID. SPL-Loc still outperforms previous methods, which demonstrates the effectiveness.

\begin{table}[!t]\scriptsize 
\caption{SOTA comparison on THUMOS-14. AVG is the average mAP at IoU thresholds 0.3:0.1:0.7. Manual denotes manual labels from \cite{8}. Uniform denotes labels from \cite{26}.}
\renewcommand{\arraystretch}{1}
\centering
\begin{tabular}{c|c|ccccc|c}
\toprule
\multirow {2}*{Supervision} & \multirow {2}*{Method} & \multicolumn {5}{c|}{mAP@IoU(\%)} & \multirow {2}*{AVG} \\
& & 0.3 & 0.4 & 0.5 & 0.6 & 0.7 & \\
\hline
\multirow {3}*{Video-level} & FTCL \cite{27} & 55.2 & 45.2 & 35.6 & 23.7 & 12.2 & 34.4 \\
& ASM-Loc \cite{28} & 57.1 & 46.8 & 36.6 & 25.2 & 13.4 & 35.8 \\
& RSKP \cite{29} & 55.8 & 47.5 & 38.2 & 25.4 & 12.5 & 35.9 \\
\hline
\multirow {3}*{Point-level} & SF-Net \cite{8} & 53.2 & 40.7 & 29.3 & 18.4 & 9.6 & 30.2 \\
\multirow {3}*{(Manual)} & Ju et al. \cite{9} & 58.1 & 46.4 & 34.5 & 21.8 & 11.9 & 34.5 \\
& LACP \cite{10} & 63.3 & 55.2 & 43.9 & 33.3 & 20.8 & 43.3 \\
& \textbf{Ours} & \textbf{64.4} & \textbf{56.7} & \textbf{45.3} & \textbf{35.2} & \textbf{22.3} & \textbf{44.8} \\
\hline
\multirow {3}*{Point-level} & SF-Net \cite{8} & 52.8 & 42.2 & 30.5 & 20.6 & 12.0 & 31.6 \\
\multirow {3}*{(Uniform)} & Ju et al. \cite{9} & 58.2 & 47.1 & 35.9 & 23.0 & 12.8 & 35.4 \\
& LACP \cite{10} & 64.6 & 56.5 & 45.3 & 34.5 & 21.8 & 44.5 \\
& \textbf{Ours} & \textbf{66.9} & \textbf{58.4} & \textbf{46.8} & \textbf{36.3} & \textbf{22.9} & \textbf{46.3} \\
\bottomrule
\end{tabular}
\end{table}

\begin{table}[!t]\scriptsize 
\caption{SOTA comparison on GTEA and BEOID. AVG is the average mAP at IoU thresholds 0.1:0.1:0.7.}
\centering
\begin{tabular}{c|c|cccc|c}
\toprule
\multirow {2}*{Dataset} & \multirow {2}*{Method} & \multicolumn {4}{c|}{mAP@IoU(\%)} & \multirow {2}*{AVG} \\
& & 0.1 & 0.3 & 0.5 & 0.7 & \\
\hline
\multirow {4}*{GTEA} & SF-Net \cite{8} & 58.0 & 37.9 & 19.3 & 11.9 & 31.0 \\
& Ju et al. \cite{9} & 59.7 & 38.3 & 21.9 & 18.1 & 33.7 \\
& LACP \cite{10} & 63.9 & 55.7 & 33.9 & 20.8 & 43.5 \\	
& \textbf{Ours} & \textbf{64.8} & \textbf{56.9} & \textbf{35.5} & \textbf{21.4} & \textbf{45.1} \\
\hline
\multirow {4}*{BEOID} & SF-Net \cite{8} & 62.9 & 40.6 & 16.7 & 3.5 & 30.9 \\
& Ju et al. \cite{9} & 63.2 & 46.8 & 20.9 & 5.8	 & 34.9 \\
& LACP \cite{10} & 76.9 & 61.4 & 42.7 & 25.1 & 51.8 \\
& \textbf{Ours} & \textbf{78.2} & \textbf{62.4} & \textbf{44.0} & \textbf{26.2} & \textbf{53.1} \\
\bottomrule
\end{tabular}
\end{table}

\clearpage
\balance


\begin{thebibliography}{34}

\bibitem{2}D.~Zhang, C.~Huang, C.~Liu, and Y.~Xu, ``Weakly supervised video anomaly detection via transformer-enabled temporal relation learning,'' {\em IEEE Signal Process. Lett.}, vol. 29, pp. 1197–1201, 2022.

\bibitem{3}W.~Ruan, Y.~Tao, L.~Ruan, X.~Shu, and Y.~Qiao, ``Temporal weighting appearance-aligned network for nighttime video retrieval,'' {\em IEEE Signal Process. Lett.}, vol. 29, pp. 2008–2012, 2022.

\bibitem{4}T. Lin, X. Zhao, H. Su, C. Wang, and M. Yang, “BSN: Boundary sensitive network for temporal action proposal generation,'' in {\em Proc. Eur. Conf. Comput. Vision}, 2018, pp. 3-19. 

\bibitem{5}X.~Qin, Y.~Ge, H.~Yu, F.~Chen, and D.~Yang, ``Spatial enhancement and temporal constraint for weakly supervised action localization,'' {\em IEEE Signal Process. Lett.}, vol. 27, pp. 1520–1524, 2020.

\bibitem{6}D.~Liu, T.~Jiang, and Y.~Wang, ``Completeness modeling and context separation for weakly supervised temporal action localization,'' in {\em Proc. IEEE Conf. Comput. Vision Pattern Recognit.}, 2019, pp. 1298-1307. 

\bibitem{7}P.~Lee, Y.~Uh, and H.~Byun, ``Background suppression network for weakly-supervised temporal action localization,'' in {\em Proc. AAAI Conf. Artif. Intell.}, vol.~34, no.~7, pp. 11320-11327, 2020. 

\bibitem{8}F.~Ma, L.~Zhu, Y.~Yang, S.~Zha, G.~Kundu, M.~Feiszli, and Z.~Shou, ``Sf-net: Single-
 supervision for temporal action localization,'' in {\em Proc. Eur. Conf. Comput. Vision}, 2020, pp. 420-437.

\bibitem{9}C.~Ju, P.~Zhao, S.~Chen, Y.~Zhang, Y.~Wang, and Q.~Tian, ``Divide and conquer for single-frame temporal action localization,'' in {\em Proc. IEEE Int. Conf. Comput. Vision}, 2021, pp. 13455-13464. 

\bibitem{10}P.~Lee and H.~Byun, ``Learning action completeness from points for weakly-supervised temporal action localization,'' in {\em Proc. IEEE Int. Conf. Comput. Vision}, 2021, pp. 13648-13657.  

\bibitem{14}H.-M. Yang, X.-Y. Zhang, F.~Yin, and C.-L. Liu, ``Robust classification with convolutional prototype learning,'' in {\em Proc. IEEE Conf. Comput. Vision Pattern Recognit.}, 2018, pp. 3474-3482.

\bibitem{15}D.~Lee, S.~Lee, and H.~Yu, ``Learnable dynamic temporal pooling for time series classification,'' in {\em Proc. AAAI Conf. Artif. Intell.}, vol.~35, no.~9, pp. 8288-8296, 2021.

\bibitem{16}S.~Paul, S.~Roy, and A.~K. Roy-Chowdhury, ``W-talc: Weakly-supervised temporal activity localization and classification,'' in {\em Proc. Eur. Conf. Comput. Vision}, 2018, pp. 563-579.

\bibitem{17}T.-Y. Lin, P.~Goyal, R.~Girshick, K.~He, and P.~Doll{\'a}r, ``Focal loss for dense object detection,'' in {\em Proc. IEEE Int. Conf. Comput. Vision}, 2017, pp. 2980-2988.   

\bibitem{18}L.~Huang, Y.~Huang, W.~Ouyang, and L.~Wang, ``Relational prototypical network for weakly supervised temporal action localization,'' in {\em Proc. AAAI Conf. Artif. Intell.}, vol.~34, no.~7, pp. 11053-11060, 2020.

\bibitem{19}M.~Cuturi and M.~Blondel, ``Soft-dtw: a differentiable loss function for time-series,'' in {\em Proc. Int. Conf. Mach. Learn.}, 2017, pp. 894-903.   

\bibitem{multi-threshold-TCAS}P.~Lee, J.~Wang, Y.~Lu, and H.~Byun, ``Weakly-supervised temporal action localization by uncertainty modeling,'' in {\em Proc. AAAI Conf. Artif. Intell.}, vol.~35, no.~3, pp. 1854-1862, 2021.

\bibitem{score}Z.~Shou, H.~Gao, L.~Zhang, K.~Miyazawa, and S.-F. Chang, ``Autoloc: Weakly-supervised temporal action localization in untrimmed videos,'' in {\em Proc. Eur. Conf. Comput. Vision}, 2018, pp. 154-171.

\bibitem{NMS}N. Bodla, B. Singh, R. Chellappa, and L. S. Davis, ``Soft-NMS-improving object detection with one line of code,'' in {\em Proc. IEEE Conf. Comput. Vision Pattern Recognit.}, 2017, pp. 5561-5569.

\bibitem{22}Y.-G. Jiang, J. Liu, A. Roshan Zamir, G. Toderici, I. Laptev, M. Shah, and R. Sukthankar, “Thumos challenge: Action recognition with a large number of classes,” 2014. [Online]. Available: http://crcv.ucf.edu/THUMOS14/

\bibitem{23}P.~Lei and S.~Todorovic, ``Temporal deformable residual networks for action segmentation in videos,'' in {\em Proc. IEEE Conf. Comput. Vision Pattern Recognit.}, 2018, pp. 6742-6751.

\bibitem{24}D.~Damen, T.~Leelasawassuk, O.~Haines, A.~Calway, and W.~W. Mayol-Cuevas, ``You-do, i-learn: Discovering task relevant objects and their modes of interaction from multi-user egocentric video.'' in {\em Proc. Brit. Mach. Vision Conf.}, vol.~2, p.~3, 2014.

\bibitem{25}J.~Carreira and A.~Zisserman, ``Quo vadis, action recognition? a new model and the kinetics dataset,'' in {\em Proc. IEEE Conf. Comput. Vision Pattern Recognit.}, 2017, pp. 6299-6308. 

\bibitem{26}D.~Moltisanti, S.~Fidler, and D.~Damen, ``Action recognition from single timestamp supervision in untrimmed videos,'' in {\em Proc. IEEE Conf. Comput. Vision Pattern Recognit.}, 2019, pp. 9915-9924.

\bibitem{27}J.~Gao, M.~Chen, and C.~Xu, ``Fine-grained temporal contrastive learning for weakly-supervised temporal action localization,'' in {\em Proc. IEEE Conf. Comput. Vision Pattern Recognit.}, 2022, pp. 19999-20009.

\bibitem{28}B.~He, X.~Yang, L.~Kang, Z.~Cheng, X.~Zhou, and A.~Shrivastava, ``Asm-loc: Action-aware segment modeling for weakly-supervised temporal action localization,'' in {\em Proc. IEEE Conf. Comput. Vision Pattern Recognit.}, 2022, pp. 13925-13935.

\bibitem{29}L.~Huang, L.~Wang, and H.~Li, ``Weakly supervised temporal action localization via representative snippet knowledge propagation,'' in {\em Proc. IEEE Conf. Comput. Vision Pattern Recognit.}, 2022, pp. 3272-3281.

\end{thebibliography}
\end{document}